\documentclass[journal]{IEEEtran}
\usepackage{graphicx,amsmath,amssymb,booktabs,array,url,xcolor,multirow}
\usepackage[caption=false,font=footnotesize]{subfig}
\usepackage[hidelinks]{hyperref}
\graphicspath{{figs/}}
\usepackage{tikz}
\usepackage{pgfplots}
\pgfplotsset{compat=1.18}
\usepackage{xcolor}

\begin{document}
\title{Explanation Stability of Test-Time Adaptation \\ in Computational Pathology: A Large-Scale Benchmark}
\author{
R.~G.~Bahumanya,
Harshith~V.~M.,
Shreyank~N.~Gowda,
and Anala~M.~R.%
\thanks{Manuscript submitted 2026. This work was did not receive any external funding.}
\thanks{R.~G.~Bahumanya, Harshith~V.~M., and Anala~M.~R. are with the Department of Computer Science and Engineering, R.V. College of Engineering, Bengaluru, India. Emails: \texttt{bahumanyarg.ci25@rvce.edu.in}, \texttt{harshithvm.ci25@rvce.edu.in}, \texttt{analamr@rvce.edu.in}.}
\thanks{Shreyank~N.~Gowda is with the School of Computer Science, University of Nottingham, Nottingham, United Kingdom. Email: \texttt{shreyank.narayanagowda@nottingham.ac.uk}.}
\thanks{R.~G.~Bahumanya and Harshith~V.~M. contributed equally.}
\thanks{Corresponding author: Anala~M.~R. \texttt{analamr@rvce.edu.in}.}
}

\markboth{IEEE Transactions on Medical Imaging,~Vol.~XX, No.~X, 2026}{Author \MakeLowercase{\textit{et al.}}: TTA Reshapes Explanations in Computational Pathology}

\maketitle

\begin{abstract}
Test-time adaptation (TTA) has become a practical way to adapt deployed models to unlabeled target data, a setting that is especially relevant in computational pathology where staining, scanner, and cohort shifts are routine. While most TTA methods are evaluated by their effect on accuracy, clinical use also depends on whether the model's explanations remain reliable after adaptation. In this paper, we take a closer look at this largely unmeasured effect. We study explanation stability under TTA across two histopathology benchmarks, Camelyon17 and NCT-CRC-HE, using five architectures ranging from convolutional networks to vision transformers and a pathology foundation model, seventeen TTA methods, and four attribution families. Across 2{,}958 adaptation runs, we observe a clear and systematic pattern: TTA methods differ sharply in how much they move model explanations, with frozen-backbone methods leaving attributions almost unchanged and continual methods such as CoTTA and RoTTA causing the largest drift. This effect is not uniform. Convolutional networks are substantially more sensitive than transformer and foundation-model backbones, and explanation drift increases with adaptation strength while remaining largely insensitive to batch size. Surprisingly, explanation stability is only weakly coupled to adaptation quality. Some methods preserve explanations almost perfectly while degrading calibration or accuracy, producing silent failures that would be missed by accuracy-only or explanation-only evaluation. These findings show that explanation stability is a distinct reliability axis for TTA in computational pathology. We release the metric, protocol, and full benchmark to support future work on adaptation methods that are not only accurate, but also stable and clinically auditable. Code: \url{https://github.com/bahumanyarg11/tta-explanation-stability-pipeline}
\end{abstract}


\section{Introduction}
\IEEEPARstart{D}{eep} learning has become a central tool in computational pathology, enabling tumour detection, tissue classification, and large-scale analysis of digitised histology. Yet these models remain sensitive to domain shift. A model trained on one cohort, scanner, staining protocol, or hospital site can degrade when deployed elsewhere~\cite{bandi2019camelyon17,koh2021wilds}. This problem is especially acute in pathology, where acquiring labelled target data for every deployment setting is expensive, slow, and often impractical.

\emph{Test-time adaptation} (TTA) offers a practical response to this setting. Rather than retraining a model with labelled target data, TTA updates the model or its prediction rule online using only the unlabelled target stream. Existing methods include batch-normalisation statistic recalibration~\cite{schneider2020bn}, entropy minimisation~\cite{wang2021tent}, continual teacher--student adaptation~\cite{wang2022cotta,yuan2023rotta}, and reliability-aware objectives~\cite{niu2023sar,zhang2025come}. These methods are usually judged by whether they improve target-domain accuracy.

Accuracy alone is not enough for clinical use. A pathologist or clinical auditor may also ask where the model looked. Post-hoc explanations such as Grad-CAM~\cite{selvaraju2017gradcam} and Integrated Gradients~\cite{sundararajan2017ig} are commonly used to inspect whether a model attends to plausible tissue regions. Although such explanations have known limitations~\cite{adebayo2018sanity,rudin2019stop}, they remain part of the practical trust pipeline around medical AI systems. This raises a simple question. When TTA changes a model at deployment time, does it also change the explanations that clinicians may inspect?

This question matters because explanation stability is not guaranteed by accuracy. A method may preserve accuracy while moving attention to different tissue regions. A method may also preserve explanations while degrading calibration or discrimination. In both cases, standard TTA evaluation can miss relevant behaviour. Accuracy, calibration, and explanation stability should therefore be treated as distinct axes of reliability.

In this work, we study explanation stability under TTA at scale. We evaluate 17 TTA methods across five architectures, ResNet-50, EfficientNet-B3, Swin-T, ViT-B/16, and Phikon-v2, on two histopathology benchmarks. Camelyon17 captures multi-hospital domain shift, while NCT-CRC-HE provides a nine-class colorectal tissue classification task. Across 2{,}958 adaptation runs, we measure discrimination, calibration, and explanation movement using four attribution families and a bounded Explanation Stability Index (ESI).

Our study leads to three main findings. First, TTA methods differ systematically in how much they perturb explanations. Frozen-backbone methods leave attributions nearly unchanged, while continual methods such as CoTTA and RoTTA produce the largest shifts. Second, explanation drift depends strongly on architecture and task difficulty. Convolutional networks are more vulnerable than transformer and foundation-model backbones, and the harder multi-class setting induces larger drift. Third, explanation stability is largely decoupled from adaptation quality. Some methods preserve explanations almost perfectly while degrading calibration or accuracy, creating failures that would be missed by accuracy-only or explanation-only evaluation.

Our contributions are as follows.
\begin{enumerate}
    \item \textbf{Explanation stability under TTA.} We formalise explanation stability as a reliability axis for test-time adaptation and introduce ESI, an equal-weighted bounded index combining complementary similarity and divergence measures.
    \item \textbf{Large-scale pathology benchmark.} We evaluate 17 TTA methods across five architectures, two histopathology datasets, multiple domains, and repeated seeds, yielding 2{,}958 adaptation runs with paired accuracy, calibration, and explanation measurements.
    \item \textbf{Empirical findings and deployment implications.} We show that explanation drift is method-dependent, architecture-dependent, and shift-dependent. We also show that drift increases with adaptation strength, and that explanation stability does not substitute for calibration or accuracy monitoring.
\end{enumerate}

\section{Related Work}

\noindent\textbf{Test-time adaptation.}
Test-time adaptation (TTA) adapts a source model to an unlabelled target stream without access to target labels~\cite{liang2024ttasurvey}. Existing methods differ in what they update and how much of the model they modify. Normalisation-based methods recalibrate batch-normalisation statistics on target batches~\cite{schneider2020bn}. Entropy-based methods update affine normalisation parameters by minimizing prediction entropy~\cite{wang2021tent}, often with anti-forgetting terms or reliable-sample filtering~\cite{niu2022eata,niu2023sar}. Continual methods use teacher--student updates, stochastic restoration, or memory mechanisms to adapt over a stream~\cite{wang2022cotta,yuan2023rotta,gong2023sotta}. Other approaches use prototype adaptation, graph-based label propagation, weight ensembling, adapter injection, or recent calibration-aware objectives~\cite{iwasawa2021t3a,boudiaf2022lame,marsden2024roid,liu2024vida,lee2024deyo,zhang2025come,han2025rem}. These methods are usually evaluated by target-domain accuracy, with calibration reported less often. Their effect on post-hoc explanations is largely unmeasured. Our work does not propose a new adaptation rule. Instead, it evaluates how existing TTA methods change model explanations.

\noindent\textbf{Explainability and explanation fragility.}
Post-hoc attribution methods are widely used to inspect image classifiers. CAM-based methods such as Grad-CAM, Grad-CAM++, and Score-CAM produce spatial heatmaps from activation and gradient information~\cite{selvaraju2017gradcam,chattopadhay2018gradcampp,wang2020scorecam}. Integrated Gradients provides an axiomatic gradient-based attribution method~\cite{sundararajan2017ig}. Transformer models are often analysed with attention rollout and related attention aggregation methods~\cite{abnar2020rollout}. At the same time, explanation methods are known to be fragile. Saliency maps can be sensitive to model randomisation, input perturbation, and implementation choices~\cite{adebayo2018sanity}. This has motivated faithfulness tests such as insertion and deletion curves~\cite{petsiuk2018rise}. Existing work studies explanation fragility under perturbations, randomisation, or masking. We instead study explanation change caused by online model adaptation.

\noindent\textbf{Calibration and trustworthy medical AI.}
Accuracy is not the only requirement for safe medical AI. Modern neural networks are often miscalibrated, meaning that their confidence scores do not reliably match empirical correctness~\cite{guo2017calibration}. Calibration is therefore a distinct reliability axis, especially when predictions may support clinical triage or quality assurance. Medical-imaging benchmarks also show that distribution shift can affect models differently across sites, scanners, and cohorts~\cite{koh2021wilds,bandi2019camelyon17}. Our study treats explanation stability as a third axis beside accuracy and calibration. We show that it is not reducible to either one, since a TTA method can preserve explanations while degrading calibration, or move explanations without a corresponding accuracy change.

\section{Methods}
\label{sec:methods}

\subsection{Problem setup}
\label{sec:problem_setup}

\noindent
Let $f_\theta$ denote a source model trained before deployment. A TTA method $\mathcal{A}$ uses an unlabelled target stream to produce an adapted model $f_{\theta'}$, or an adapted prediction rule, without access to target labels. For an input image $x$ and an attribution operator $\phi$, we define the pre-adaptation and post-adaptation saliency maps as
\begin{equation}
    S_{\mathrm{pre}} = \phi(f_\theta, x),
    \qquad
    S_{\mathrm{post}} = \phi(f_{\theta'}, x).
\end{equation}
We study three coupled effects of adaptation. These are the change in explanation from $S_{\mathrm{pre}}$ to $S_{\mathrm{post}}$, the change in discriminative performance measured by $\Delta\mathrm{AUC}$, and the change in calibration measured by $\Delta\mathrm{ECE}$.

\subsection{Explanation Stability Index}
\label{sec:esi}

\noindent
We quantify explanation movement with the Explanation Stability Index (ESI). Each saliency map is min-max normalised to $[0,1]$ and resized to a common $224 \times 224$ grid. For divergence-based terms, the map is also converted to a probability distribution with additive smoothing $\varepsilon=10^{-8}$ to avoid unstable values in near-zero background regions.

\noindent
ESI combines six agreement measures. Four are similarity terms, namely structural similarity SSIM~\cite{wang2004ssim}, cosine similarity, rescaled Spearman rank correlation $\rho'=\tfrac{1}{2}(\rho+1)$, and top-20\% region-of-interest IoU. Two are divergence terms, namely symmetric KL divergence and Wasserstein-1 distance, also referred to as EMD. The symmetric KL term is
\begin{equation}
    \mathrm{KL}_{\mathrm{sym}}(p,q)
    =
    \tfrac{1}{2}
    \left[
    \mathrm{KL}(p\Vert q) + \mathrm{KL}(q\Vert p)
    \right].
\end{equation}
Both divergence terms are normalised to $[0,1]$ using fixed caps and then complemented so that larger values always denote greater stability.

\noindent
We use equal weights to avoid tuning the index toward any particular method ordering:
\begin{equation}
\mathrm{ESI}
=
\tfrac{1}{6}
\left(
\mathrm{SSIM}
+
\cos
+
\rho'
+
\mathrm{IoU}
+
(1-\mathrm{KL}_{n})
+
(1-\mathrm{EMD}_{n})
\right).
\end{equation}
All terms lie in $[0,1]$, so ESI is bounded in $[0,1]$, with $\mathrm{ESI}=1$ for an unchanged explanation.

\subsection{Datasets and models}
\label{sec:datasets_models}

\noindent
We use two public histopathology benchmarks with natural domain shift rather than synthetic corruption. \textbf{Camelyon17} contains lymph-node tumour and normal patches from five hospital centres~\cite{bandi2019camelyon17,koh2021wilds}. The source model is trained on PatchCamelyon~\cite{veeling2018pcam} and adapted and evaluated separately on each Camelyon17 centre. Each centre contains 59k to 147k patches, with 493 evaluation runs per centre after subsampling. \textbf{NCT-CRC-HE} is a nine-class colorectal tissue classification benchmark~\cite{kather2019nct}. We train on the 100k-image cohort and adapt to the independent 7k-image cohort, which provides a multi-class cohort-shift setting.

\noindent
We evaluate five architectures that span convolutional, transformer, and foundation-model backbones. The convolutional models are ResNet-50~\cite{he2016resnet} and EfficientNet-B3~\cite{tan2019efficientnet}. The transformer models are Swin-T~\cite{liu2021swin} and ViT-B/16~\cite{dosovitskiy2021vit}. The pathology foundation model is Phikon-v2~\cite{filiot2023phikon}. For transformer and foundation models, the backbone is frozen except for the final normalisation blocks and the linear head, following common practice for adapting large models.

\subsection{Adaptation and attribution methods}
\label{sec:adaptation_attribution}

\noindent
Table~\ref{tab:tax} lists the seventeen TTA methods grouped by mechanism. Explanations are computed with four attribution families. Integrated Gradients~\cite{sundararajan2017ig} is our primary explainer because it is architecture-agnostic and avoids target-layer selection. We also use Grad-CAM~\cite{selvaraju2017gradcam}, Grad-CAM++~\cite{chattopadhay2018gradcampp}, and attention rollout~\cite{abnar2020rollout}. Faithfulness is evaluated with insertion and deletion AUC~\cite{petsiuk2018rise}.

\begin{table*}[t]
\caption{The seventeen evaluated TTA methods grouped by mechanistic family.}
\label{tab:tax}
\centering
\footnotesize
\begin{tabular}{@{}l l l@{}}
\toprule
Family & Methods & Representative venues \\
\midrule
Baseline & No-TTA & -- \\
Normalisation & BN-Adapt~\cite{schneider2020bn} & NeurIPS 2020 \\
Entropy & Tent~\cite{wang2021tent}, EATA~\cite{niu2022eata}, DeYO~\cite{lee2024deyo}, ROID~\cite{marsden2024roid}, COME~\cite{zhang2025come}, REM~\cite{han2025rem} & ICLR 2021 to 2025 \\
Sharpness & SAR~\cite{niu2023sar}, SoTTA~\cite{gong2023sotta} & ICLR 2023 \\
Continual & CoTTA~\cite{wang2022cotta}, RoTTA~\cite{yuan2023rotta}, RMemSafe~\cite{rmemsafe2026} & CVPR 2022 to 2026 \\
Prototype & T3A~\cite{iwasawa2021t3a} & NeurIPS 2021 \\
Gradient-free & LAME~\cite{boudiaf2022lame}, SICL~\cite{sicl2025} & CVPR 2022, WACV 2026 \\
Adapter & ViDA~\cite{liu2024vida} & ICLR 2024 \\
\bottomrule
\end{tabular}
\end{table*}

\noindent
We briefly summarise the mechanisms of the evaluated TTA families. Let $p=\sigma(f_\theta(x))$ denote the softmax prediction and let
\begin{equation}
    H(p) = -\sum_k p_k \log p_k
\end{equation}
denote prediction entropy.

\noindent
\textbf{Normalisation methods.}
BN-Adapt replaces source batch-normalisation statistics $(\mu_s,\sigma_s^2)$ with target-batch statistics $(\mu_t,\sigma_t^2)$ while leaving the learned weights fixed~\cite{schneider2020bn}.

\noindent
\textbf{Entropy methods.}
Tent minimises prediction entropy over affine normalisation parameters~\cite{wang2021tent}. EATA adds sample filtering and a Fisher anti-forgetting penalty~\cite{niu2022eata}. DeYO weights samples using a patch-shuffling disentanglement score~\cite{lee2024deyo}. ROID combines certainty and diversity weighting with weight ensembling and prior correction~\cite{marsden2024roid}. COME uses an evidential entropy objective to reduce overconfidence~\cite{zhang2025come}. REM ranks entropy across progressively masked views~\cite{han2025rem}.

\noindent
\textbf{Sharpness and continual methods.}
SAR applies sharpness-aware minimisation with reliable-sample selection~\cite{niu2023sar}. SoTTA combines a high-confidence uniform-class memory with entropy and sharpness minimisation~\cite{gong2023sotta}. CoTTA uses a weight-averaged teacher with augmentation consistency and stochastic weight restoration~\cite{wang2022cotta}. RoTTA adds category-balanced memory with timeliness-aware and uncertainty-aware reweighting~\cite{yuan2023rotta}. RMemSafe gates source anchoring by predictive reliability~\cite{rmemsafe2026}.

\noindent
\textbf{Prototype, gradient-free, and adapter methods.}
T3A updates class templates from confident test features without gradient updates~\cite{iwasawa2021t3a}. LAME performs Laplacian-regularised label smoothing over a feature-similarity graph~\cite{boudiaf2022lame}. SICL rescales confidence using style-invariance~\cite{sicl2025}. ViDA injects low-rank and high-rank adapters updated with a teacher~\cite{liu2024vida}.

\noindent
These families differ in how much of the network they modify. Normalisation and entropy methods update parameters that affect the feature hierarchy. Continual methods add teacher drift and memory effects. Prototype and gradient-free methods leave the backbone unchanged, which means their attribution maps can remain unchanged even when predictions or calibration change.

\subsection{Attribution operators}
\label{sec:attribution_methods}

\noindent
Integrated Gradients attributes a prediction to input dimensions by integrating gradients along a straight path from a baseline input $x'$ to the image $x$~\cite{sundararajan2017ig}:
\begin{equation}
    \mathrm{IG}_i(x)
    =
    (x_i-x_i')
    \int_0^1
    \partial_i f(x' + \alpha(x-x'))\,d\alpha .
\end{equation}
We use Integrated Gradients as the primary explainer because it applies to all evaluated architectures without choosing a target layer.

\noindent
Grad-CAM and Grad-CAM++ compute class-discriminative heatmaps by weighting activations from a target convolutional or transformer block using first-order or higher-order gradients~\cite{selvaraju2017gradcam,chattopadhay2018gradcampp}. Attention rollout multiplies layer-wise attention matrices to obtain transformer-native attribution maps~\cite{abnar2020rollout}. Using multiple attribution families lets us test whether the ranking of TTA methods is an explainer artifact.

\subsection{Evaluation protocol and implementation}
\label{sec:evaluation_protocol}

\noindent
For each dataset, model, TTA method, domain, and seed, we adapt on the full target stream. We then compute AUC, ECE~\cite{guo2017calibration}, and ESI on a fixed random subset that is identical across methods within a domain. This preserves paired comparisons. We repeat the evaluation over five prime-number seeds, and over seven seeds for the convolutional models. The full benchmark contains 2{,}958 adaptation runs.

\noindent
Source models are trained with AdamW under a cosine learning-rate schedule, with early stopping based on validation AUC. For convolutional models, the classifier heads are trainable. For transformer and foundation backbones, only the final normalisation blocks and the linear head are trainable. TTA hyperparameters follow the authors' recommended defaults, and adaptation uses one gradient step per batch except in the step-ablation.

\noindent
All test-time inference is performed in single precision. As a correctness check, the No-TTA condition produces numerically identical pre-adaptation and post-adaptation logits, with maximum $|\Delta\mathrm{AUC}|=0$ across all No-TTA runs. Each Camelyon17 evaluation uses a fixed 15{,}000-patch stream per hospital centre, identical across methods and seeds. For NCT-CRC-HE, we use the full independent cohort. Saliency statistics are accumulated over a fixed patch set for each run.

\noindent
Statistical analyses use \texttt{statsmodels} for linear mixed-effects models, \texttt{scipy} and \texttt{scikit-posthocs} for Friedman and Nemenyi tests, and 2{,}000-sample bootstrap resampling for confidence intervals. We model $\Delta\mathrm{ECE}$ using ESI and $\Delta\mathrm{AUC}$ as fixed effects with random intercepts for model and dataset. We also report the partial Spearman correlation between ESI and $\Delta\mathrm{ECE}$ after controlling for $\Delta\mathrm{AUC}$. Method comparisons use the Friedman test over matched blocks and a Nemenyi critical-difference analysis~\cite{demsar2006}.

\section{Results}
\label{sec:results}

\subsection{TTA methods form a clear hierarchy of explanation stability}
\label{sec:results_main}

\begin{figure}[t]
\centering
\includegraphics[width=\columnwidth]{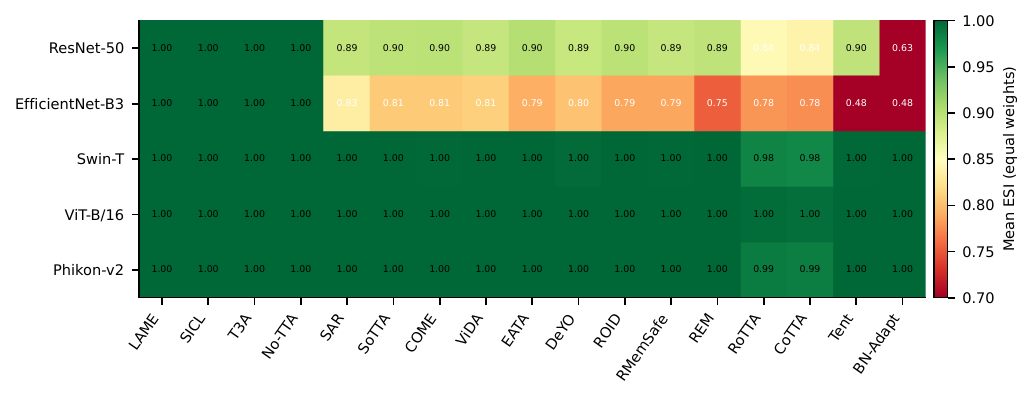}
\caption{Mean ESI across architectures and TTA methods. Higher values indicate greater explanation stability. Frozen-backbone methods preserve explanations, while continual methods and BN-based adaptation are the primary disruptors.}
\label{fig:heatmap}
\end{figure}

\begin{figure}[t]
\centering
\includegraphics[width=0.82\columnwidth]{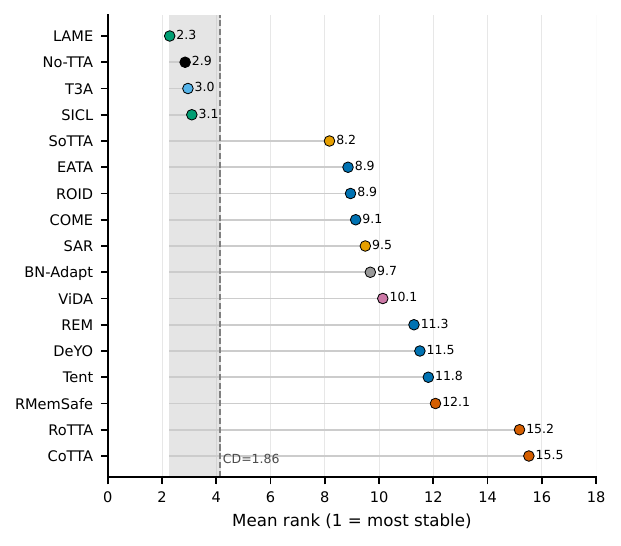}
\caption{Nemenyi mean-rank ranking by ESI. Rank $1$ denotes the most stable method. The critical difference is $1.86$, showing that continual methods separate clearly from the most stable group.}
\label{fig:cd}
\end{figure}

\noindent
Fig.~\ref{fig:heatmap} and Table~\ref{tab:main} summarize the main benchmark results across 2{,}958 adaptation runs. TTA methods differ strongly in how much they perturb explanations. The Friedman test rejects equality across methods with $\chi^2_{16}=1828$, $p<10^{-300}$, over $174$ matched blocks. The Nemenyi analysis gives a critical difference of $1.86$ and separates the most stable methods from the strongest disruptors, as shown in Fig.~\ref{fig:cd}.

\noindent
The ordering is mechanistically interpretable. Methods that leave the backbone fixed, including No-TTA, LAME, T3A, and SICL, preserve explanations almost perfectly with $\mathrm{ESI}=1.000$. Continual methods are the most consistent disruptors. CoTTA and RoTTA obtain the worst mean ranks, $15.5$ and $15.2$, because they perturb explanations across architectures. BN-Adapt and Tent have low pooled ESI values, $0.786$ and $0.849$, but their ranks are less extreme because their disruption is concentrated on convolutional models.

\begin{table}[t]
\caption{Main pooled results across five architectures and two datasets. ESI denotes explanation stability, where $1$ means unchanged. Rank is the Nemenyi mean rank, where lower is more stable.}
\label{tab:main}
\centering
\footnotesize
\setlength{\tabcolsep}{4pt}
\begin{tabular}{lrrrr}
\toprule
Method & $\Delta$AUC & $\Delta$ECE & ESI & Rank \\
\midrule
LAME      & +0.000 & +0.027 & 1.000 & 2.3 \\
No-TTA    & +0.000 & +0.000 & 1.000 & 2.9 \\
T3A       & -0.134 & +0.199 & 1.000 & 3.0 \\
SICL      & +0.000 & -0.001 & 1.000 & 3.1 \\
SoTTA     & +0.000 & -0.000 & 0.928 & 8.2 \\
EATA      & +0.000 & +0.000 & 0.925 & 8.9 \\
ROID      & -0.039 & +0.020 & 0.924 & 8.9 \\
COME      & +0.001 & -0.000 & 0.928 & 9.1 \\
SAR       & +0.001 & -0.000 & 0.934 & 9.5 \\
BN-Adapt  & -0.016 & +0.010 & 0.786 & 9.7 \\
ViDA      & +0.000 & +0.002 & 0.928 & 10.1 \\
REM       & -0.000 & +0.001 & 0.915 & 11.3 \\
DeYO      & +0.001 & -0.000 & 0.925 & 11.5 \\
Tent      & +0.001 & -0.000 & 0.849 & 11.8 \\
RMemSafe  & +0.000 & -0.000 & 0.922 & 12.1 \\
RoTTA     & -0.002 & -0.000 & 0.903 & 15.2 \\
CoTTA     & +0.001 & -0.000 & 0.900 & 15.5 \\
\bottomrule
\end{tabular}
\end{table}

\subsection{Architecture and shift severity control explanation drift}
\label{sec:results_architecture}

\noindent
Explanation drift is not uniform across backbones. Table~\ref{tab:arch_silent} shows that convolutional models are substantially more affected than transformer and foundation-model backbones. EfficientNet-B3 has the lowest mean ESI at $0.806$, followed by ResNet-50 at $0.898$. In contrast, Swin-T, ViT-B/16, and Phikon-v2 remain close to $1.00$. This pattern matches the adaptation mechanisms. In convolutional networks, BN-statistic and affine-parameter updates affect the feature hierarchy broadly. In the transformer and foundation-model settings, adaptation modifies only a small norm-affine subset while the backbone remains mostly frozen.

\noindent
Shift difficulty also affects drift. Across Camelyon17 hospital centres, frozen methods remain flat while disruptive methods vary with site. The harder nine-class NCT-CRC-HE task increases drift for continual methods. CoTTA changes from ESI $0.907$ on Camelyon17 to $0.868$ on NCT-CRC-HE, and RoTTA changes from $0.909$ to $0.870$. These results show that explanation stability depends on both the adaptation rule and the deployment shift.

\begin{table}[t]
\caption{Architecture-level stability and silent calibration failures. Silent failures have stable accuracy, $|\Delta\mathrm{AUC}|<0.02$, but degraded calibration, $\Delta\mathrm{ECE}>0.05$.}
\label{tab:arch_silent}
\centering
\footnotesize
\setlength{\tabcolsep}{5pt}
\begin{tabular}{lrr@{\hspace{1.5em}}lr}
\toprule
\multicolumn{3}{c}{\textbf{A. Architecture}} &
\multicolumn{2}{c}{\textbf{B. Silent failures}} \\
\cmidrule(lr){1-3}
\cmidrule(lr){4-5}
Backbone & ESI & Runs & Source & Count \\
\midrule
ResNet-50       & 0.898 & 714 & LAME & 31 \\
EfficientNet-B3 & 0.806 & 714 & T3A & 20 \\
Swin-T          & 0.997 & 510 & Other & 3 \\
ViT-B/16        & 0.999 & 510 & Total & 54 \\
Phikon-v2       & 0.998 & 510 &  &  \\
\bottomrule
\end{tabular}
\end{table}

\subsection{Adaptation changes where the model attends}
\label{sec:results_mechanism}

\begin{figure}[t]
\centering
\includegraphics[width=\columnwidth]{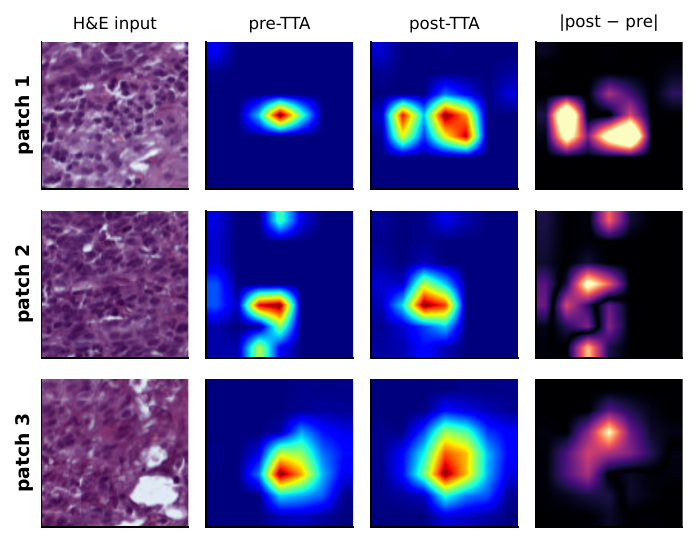}
\caption{Grad-CAM attributions before and after CoTTA on ResNet-50 for representative Camelyon17 patches. Difference maps show that continual adaptation relocates high-attention regions.}
\label{fig:qual}
\end{figure}

\noindent
The qualitative examples in Fig.~\ref{fig:qual} show that explanation drift corresponds to spatial relocation. For CoTTA on ResNet-50, high-attribution regions split, shift, or expand after adaptation. These changes occur on the same input image, so they reflect adaptation-induced changes in the model rather than changes in tissue content.

\begin{figure}[t]
\centering
\includegraphics[width=\columnwidth]{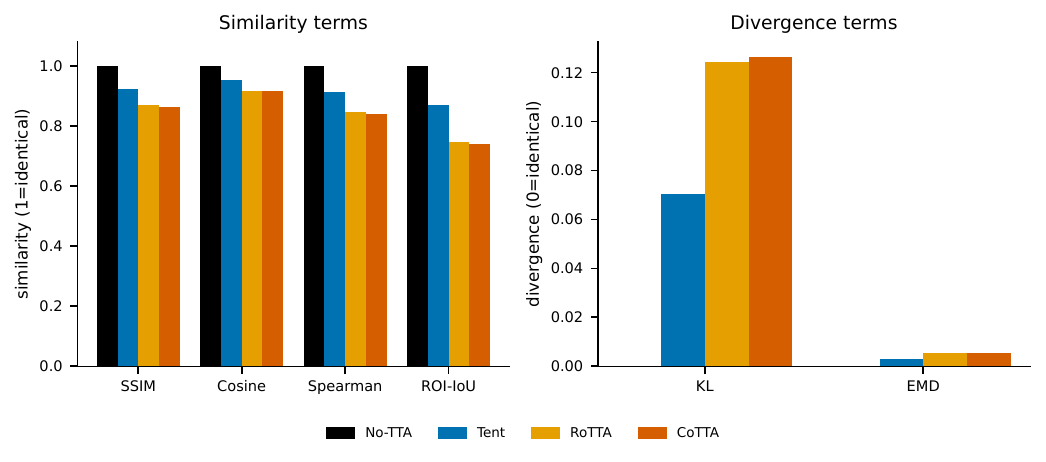}
\caption{ESI component decomposition for representative methods. Continual methods reduce region overlap and increase distributional divergence, indicating spatial relocation of attention.}
\label{fig:decomp}
\end{figure}

\noindent
The component analysis in Fig.~\ref{fig:decomp} confirms this interpretation quantitatively. Continual methods reduce top-region IoU and increase KL and EMD, indicating movement of attribution mass rather than simple heatmap rescaling. BN-Adapt shows the strongest distributional shift among the representative methods, with reduced SSIM, cosine similarity, Spearman correlation, and IoU.

\noindent
Ablations support adaptation as the driver of this effect. Increasing the number of Tent adaptation steps monotonically decreases ESI from $0.890$ to $0.883$, $0.875$, and $0.864$ for $1$, $2$, $4$, and $8$ steps. By contrast, changing batch size has little effect on ESI. Explanation drift therefore follows adaptation strength rather than batch size alone.

\subsection{Explanation stability is distinct from adaptation quality}
\label{sec:results_decoupling}

\noindent
The central reliability finding is that explanation stability does not imply safe adaptation. T3A preserves explanations perfectly with $\mathrm{ESI}=1.000$, but loses accuracy and calibration, with $\Delta\mathrm{AUC}=-0.134$ and $\Delta\mathrm{ECE}=+0.199$. LAME also preserves explanations while increasing ECE by $+0.027$. These methods leave the backbone unchanged, so attribution maps can remain stable even when predictions or confidence values degrade.

\noindent
This decoupling produces silent calibration failures. We define such a failure as stable accuracy, $|\Delta\mathrm{AUC}|<0.02$, with degraded calibration, $\Delta\mathrm{ECE}>0.05$. We observe $54$ failures, of which $51$ come from LAME or T3A. Thus, $94\%$ of silent calibration failures arise from methods that appear explanation-stable.

\begin{figure}[t]
\centering
\includegraphics[width=\columnwidth]{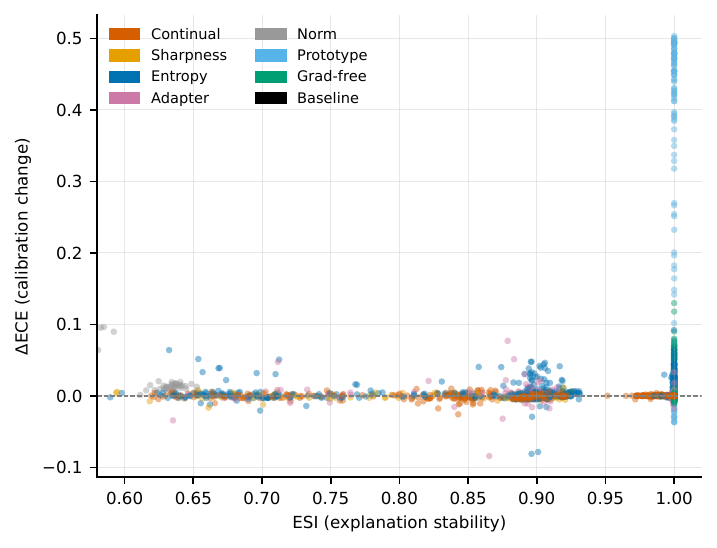}
\caption{Explanation stability versus calibration change. Stable explanations do not guarantee safe adaptation. Several methods preserve explanations while degrading calibration.}
\label{fig:esi_ece}
\end{figure}

\noindent
A mixed-effects analysis gives the same conclusion. After controlling for $\Delta\mathrm{AUC}$, the partial Spearman correlation between ESI and $\Delta\mathrm{ECE}$ is small, with $\rho=+0.089$ and $p=1.3\times10^{-6}$. The coefficient on $\Delta\mathrm{AUC}$ is much larger than the coefficient on ESI, with $-0.952$ compared with $+0.055$. Accuracy change dominates calibration change, while explanation stability remains a separate reliability axis.

\subsection{Robustness checks}
\label{sec:results_robustness}

\noindent
The explanation-stability ranking is not an artifact of one attribution method. Method rankings under Integrated Gradients correlate with Grad-CAM and Grad-CAM++ with Spearman correlations of $0.56$ and $0.59$, both with $p<0.02$. Attention rollout is evaluated only for transformer models and is used as an architecture-specific check rather than as the primary ranking signal.

\noindent
The ranking is also robust to the ESI weighting. Reweighting similarity and divergence terms over $\{0.5,0.6,0.7,0.8,0.9\}$ leaves the method ranking unchanged, with Spearman rank correlation $1.00$ between every pair of settings. Faithfulness checks using insertion and deletion AUC show that post-adaptation attributions remain usable, while saliency KL tracks accuracy degradation only moderately with $\rho=0.33$ against $|\Delta\mathrm{AUC}|$. Together, these checks support the main conclusion that accuracy, calibration, and explanation stability must be measured together.

\noindent
Table~\ref{tab:matrix} gives the complete numeric ESI matrix corresponding to Fig.~\ref{fig:heatmap}. The convolutional and transformer gap is consistent across methods. BN-Adapt and Tent strongly disrupt EfficientNet-B3 and ResNet-50, while Swin-T, ViT-B/16, and Phikon-v2 remain close to unchanged. Continual methods form the lowest-stability group on convolutional backbones but are less disruptive on transformer and foundation-model architectures.

\begin{table}[t]
\caption{Mean ESI for every architecture and TTA method. This is the numeric counterpart of Fig.~\ref{fig:heatmap}.}
\label{tab:matrix}
\centering
\footnotesize
\setlength{\tabcolsep}{3.2pt}
\begin{tabular}{lccccc}
\toprule
Method & ResNet & Eff.-B3 & Swin & ViT & Phikon \\
\midrule
LAME      & 1.000 & 1.000 & 1.000 & 1.000 & 1.000 \\
No-TTA    & 1.000 & 1.000 & 1.000 & 1.000 & 1.000 \\
T3A       & 1.000 & 1.000 & 1.000 & 1.000 & 1.000 \\
SICL      & 1.000 & 1.000 & 1.000 & 1.000 & 1.000 \\
SoTTA     & 0.897 & 0.808 & 0.999 & 1.000 & 1.000 \\
EATA      & 0.901 & 0.792 & 0.999 & 1.000 & 1.000 \\
ROID      & 0.898 & 0.787 & 0.999 & 1.000 & 1.000 \\
COME      & 0.897 & 0.807 & 0.999 & 1.000 & 1.000 \\
SAR       & 0.893 & 0.834 & 0.999 & 1.000 & 1.000 \\
BN-Adapt  & 0.633 & 0.479 & 1.000 & 1.000 & 1.000 \\
ViDA      & 0.891 & 0.810 & 0.999 & 1.000 & 1.000 \\
REM       & 0.895 & 0.754 & 0.999 & 1.000 & 1.000 \\
DeYO      & 0.890 & 0.802 & 0.997 & 1.000 & 1.000 \\
Tent      & 0.895 & 0.482 & 0.999 & 1.000 & 1.000 \\
RMemSafe  & 0.892 & 0.788 & 0.998 & 1.000 & 1.000 \\
RoTTA     & 0.845 & 0.779 & 0.982 & 0.995 & 0.986 \\
CoTTA     & 0.839 & 0.776 & 0.980 & 0.995 & 0.985 \\
\bottomrule
\end{tabular}
\end{table}

\section{Discussion}
\label{sec:discussion}

\noindent
Our experiments show that TTA can change model explanations in a systematic and measurable way. This change is not random. It depends on the adaptation method, the model architecture, the deployment shift, and the strength of adaptation. The central finding is that explanation stability cannot be inferred from accuracy or calibration alone. Accuracy, calibration, and explanation stability therefore describe distinct aspects of TTA reliability.

\subsection{Implications for deployment}
\label{sec:discussion_deployment}

\noindent
The results have two practical implications. First, method choice affects interpretability independently of accuracy. Continual methods such as CoTTA and RoTTA are the most consistent explanation disruptors, while BN-Adapt and Tent are particularly disruptive on convolutional backbones. In contrast, reliable-sample, sharpness-aware, and frozen-backbone methods tend to move attributions less. Second, explanation stability does not guarantee safe adaptation. T3A and LAME preserve attributions almost perfectly, yet both can degrade calibration, and T3A also reduces accuracy. This creates a failure mode in which the explanation appears unchanged while the predictive behaviour becomes less reliable.

\noindent
These findings suggest that explanation stability should be reported alongside accuracy and calibration when evaluating TTA in medical imaging. For deployment, an ESI-like statistic can be logged against the frozen source model as a label-free integrity check. Such a statistic should not replace calibration monitoring or performance validation. Rather, it should act as an additional signal that the adapted model may no longer be using the same visual evidence as the source model.

\subsection{Why some methods disrupt explanations more}
\label{sec:discussion_mechanism}

\noindent
The method ordering is largely explained by how much of the representation is modified. Continual methods accumulate updates over the target stream, which can gradually move the teacher and student away from the source model. BN-Adapt replaces batch-normalisation statistics, and Tent updates normalisation parameters through entropy minimisation. These changes affect features throughout convolutional networks, which explains why ResNet-50 and EfficientNet-B3 show the largest explanation drift.

\noindent
Transformer and foundation-model backbones are more stable in our benchmark. In these models, adaptation is restricted to a small set of normalisation parameters and the linear head, while most of the backbone remains frozen. This limits how much the internal representation can move and helps explain the near-perfect ESI values observed for Swin-T, ViT-B/16, and Phikon-v2. Architecture is therefore not only a source of variation. It is also a design choice that can reduce explanation drift under adaptation.

\subsection{Relation to calibration}
\label{sec:discussion_calibration}

\noindent
The decoupling between explanation stability and calibration is one of the most important findings. A simple hypothesis would be that unstable explanations cause miscalibration, or that stable explanations imply reliable confidence. Our results do not support either interpretation. In the mixed-effects analysis, calibration change is dominated by accuracy change, while the association between ESI and $\Delta\mathrm{ECE}$ is small. More importantly, the strongest calibration failures occur in methods whose explanations remain stable.

\noindent
This result changes how ESI should be interpreted. ESI is not a calibration predictor. It measures whether the visual evidence used by the adapted model remains close to that of the source model. A high ESI means that the attribution map is stable, not that the prediction is correct or well calibrated. A low ESI means that adaptation has moved the explanation, not necessarily that the adapted prediction is worse. This is why the three quantities must be measured together.

\subsection{Toward explanation-stable adaptation}
\label{sec:discussion_future}

\noindent
The benchmark also suggests a path toward more reliable TTA methods. Explanation drift increases with adaptation strength and is most visible in methods that update broad parts of the feature hierarchy. Future methods could therefore regularise explanation movement directly, constrain updates to smaller parameter subsets, or combine adaptation with a source-reference stability penalty. The goal would not be to freeze explanations completely, since some change may be appropriate under real domain shift. The goal is to prevent unnecessary relocation of attention while retaining the benefits of adaptation.

\noindent
This direction is especially relevant in computational pathology, where explanations may be inspected for tissue plausibility. A useful adaptation method should improve target-domain performance without moving attention toward irrelevant structures or away from diagnostically meaningful tissue. Our metric and protocol provide a way to measure this behaviour across methods, architectures, and shifts.

\subsection{Practical recommendations}
\label{sec:discussion_recommendations}

\noindent
For diagnostic pipelines that expose explanations to clinicians or auditors, we recommend evaluating TTA with three quantities rather than one. These are discrimination, calibration, and explanation stability. When explanation stability is important, reliable-sample and sharpness-aware methods are preferable to plain BN recalibration, plain entropy minimisation, or continual adaptation on convolutional backbones. When continual adaptation is used, the number of adaptation steps should be kept small unless a stability budget is explicitly enforced.

\noindent
We also recommend treating preserved explanations with caution. Methods such as T3A and LAME can leave saliency maps unchanged while degrading calibration. Stable explanations should therefore not be interpreted as evidence of safe adaptation. They should instead be read together with calibration and performance metrics.

\begin{table}[t]
\caption{Recommended reporting items for test-time adaptation in computational pathology.}
\label{tab:reporting}
\centering
\footnotesize
\setlength{\tabcolsep}{4pt}
\begin{tabular}{p{0.28\linewidth}p{0.60\linewidth}}
\toprule
Item & Purpose \\
\midrule
Discrimination & Report pre- and post-TTA AUC or task-specific operating-point metrics. \\
Calibration & Report ECE or reliability curves before and after adaptation. \\
Explanation stability & Report ESI or an equivalent source-referenced stability measure. \\
Architecture & State which parameters are adapted and whether the backbone is frozen. \\
Adaptation strength & Report batch size, update steps, stream mode, and reset policy. \\
Failure screening & Flag runs with stable accuracy but degraded calibration or explanation drift. \\
\bottomrule
\end{tabular}
\end{table}

\noindent
Table~\ref{tab:reporting} summarizes the reporting items that follow from our results. The key point is not that ESI should replace existing metrics, but that it should be reported beside them. A TTA method can preserve explanations while degrading calibration, or alter explanations without a large accuracy change. Reporting these quantities together makes such failures visible.

\subsection{Clinical interpretation}
\label{sec:discussion_clinical}

\noindent
In computational pathology, explanation drift should be interpreted as a change in the visual evidence used by the model, not as proof that the adapted model is clinically wrong. Some movement may be appropriate when target slides differ in stain, scanner, or tissue preparation. The concern is unmanaged movement. If adaptation shifts attribution away from tumour regions, glandular structure, lymphocyte-rich areas, or other diagnostically plausible tissue patterns, then the model may remain numerically strong while becoming harder to audit. This is why we frame ESI as an integrity measure rather than a safety certificate. It identifies cases where adaptation has changed the model's visual rationale enough to justify closer inspection. In a clinical workflow, such a signal could be used to trigger review, recalibration, or rollback to the frozen source model.

\subsection{Limitations}
\label{sec:discussion_limitations}

\noindent
This study has some limitations. First, ESI is a summary index. Although its method ordering is stable under reweighting and across attribution families, its absolute value should not be interpreted as a clinical safety score. Second, perfect stability for frozen-backbone methods is partly expected, since these methods do not modify the feature extractor. This is why we emphasize the decoupling between ESI, accuracy, and calibration rather than treating high ESI as universally desirable.

\noindent
Third, our experiments use patch-level benchmarks rather than full whole-slide inference pipelines. Whole-slide aggregation, tissue sampling, and slide-level decision thresholds may introduce additional failure modes. Finally, our evaluation is computational rather than reader-based. Future work should test whether adaptation-induced explanation drift changes pathologist trust, review time, or diagnostic decisions in prospective reader studies.

\section{Conclusion}
\label{sec:conclusion}

\noindent
We studied what test-time adaptation does to model explanations in computational pathology. Across 2{,}958 runs covering 17 methods, five architectures, and two histopathology datasets, TTA methods separate clearly by how much they perturb attributions. Explanation drift is method-dependent, architecture-dependent, and shift-dependent. It also increases with adaptation strength.
The central finding is that explanation stability is distinct from adaptation quality. Some methods move explanations substantially, while others preserve explanations almost perfectly yet still degrade calibration or accuracy. This means that accuracy, calibration, and explanation stability should be measured together when evaluating TTA for medical imaging. We release the metric, protocol, and benchmark results to support future work on adaptation methods that remain accurate, calibrated, and explanation-stable under deployment shift.

\bibliographystyle{IEEEtran}
\bibliography{paper}
\end{document}